\documentclass[a4paper]{article}
\usepackage{iwslt15,amssymb,amsmath,epsfig}
\def\reg{{\rm\ooalign{\hfil
     \raise.07ex\hbox{\scriptsize R}\hfil\crcr\mathhexbox20D}}}

\title{QCRI Machine Translation Systems for IWSLT 16}

\name{}
\name{Nadir Durrani, \hspace{1mm} Fahim Dalvi, \hspace{1mm} Hassan Sajjad, \hspace{1mm} Stephan Vogel}
\address{Qatar Computing Research Institute -- HBKU \\
{\small \tt \{ndurrani,faimaduddin,hsajjad,svogel\}@qf.org.qa}
}

\usepackage{color}

\begin{document}
\maketitle
\begin{abstract}

This paper describes QCRI's machine translation systems for the IWSLT 2016 evaluation campaign. We participated in the Arabic$\rightarrow$English and English$\rightarrow$Arabic tracks. We built both Phrase-based and Neural machine translation models, in an effort to probe whether the newly emerged NMT framework surpasses the traditional phrase-based systems in Arabic-English language pairs. We trained a very strong phrase-based system including, a big language model, the Operation Sequence Model, Neural Network Joint Model and Class-based models along with different domain adaptation techniques such as MML filtering, mixture modeling and using fine tuning over NNJM model. However, a Neural MT system, trained by stacking data from different genres through fine-tuning, and applying ensemble over 8 models, beat our very strong phrase-based system by a significant 2 BLEU points margin in Arabic$\rightarrow$English direction. We did not obtain similar gains in the other direction but were still able to outperform the phrase-based system. We also applied system combination on phrase-based and NMT outputs.

\end{abstract}

\vspace{-2mm}
\section{Introduction}

We describe QCRI's phrase-based and Neural MT systems. We participated in the Arabic-to-English and English-to-Arabic MT tracks. Our translation engines have been historically based on the phrase-based system trained using the Moses toolkit \cite{Moses:2007}, but during the course of this evaluation, we made a transition towards the newly emerged Neural MT framework \cite{bahdanau:ICLR:2015}, using Nematus, a toolkit used by the top performing team \cite{sennrich-haddow-birch:2016:WMT}, during the recent WMT campaign. 

An interesting challenge associated with the IWSLT campaign is the problem of domain adaptation. The in-domain data based on TED talks is available in very little quantity compared to the out-domain UN corpus \cite{ZiemskiJP16}, which has been found to be harmful previously when simply concatenated to the training \cite{sajjad-etal:iwslt13}. In this year's IWSLT, two additional data resources Opus subtitles \cite{LISON16.947} and the QED corpus \cite{Abdelali_2014_lrec} were introduced. The latter was also used as an official test-set. Therefore apart from exploring phrase-based versus Neural MT, we geared ourselves towards adapting our system towards TED and QED talks in this multi-domain scenario. With these goals in mind we re-explored both model weighting and data filtering techniques, in these new data settings. Below we itemize the most successful attributes of our phrase-based system:

\begin{itemize}

\item We applied MML-based data selection \cite{Axelrod_2011_emnlp} to the UN and Open Sub-title data, with the goals of filtering out harmful data. 


\item We trained OSM models \cite{durrani-schmid-fraser:2011:ACL-HLT2011} on separate corpora, and interpolated them \cite{durraniEtAl:MT-Summit2015} by optimizing perplexity on the tuning-set. We also tried this on the OSM models trained on the word classes \cite{Durrani-osm-coling14}.


\item We tried the fine-tuning method of training the NNJM model on the out-domain data and fine-tuning with the in-domain TED data \cite{Luong-Manning:iwslt15}.

\item We trained big language models using all the English mono data available from the WMT campaign and giga word corpus for Arabic.

\end{itemize}

We trained our Neural MT system using the Nematus toolkit. We used Bidirectional RNN's for the encoder, 1024 LSTM units, and a word embedding size of 500. Below we itemize what worked when training the neural MT system:

\begin{itemize}

\item We trained our baseline model on all of the UN corpus, then continued training with the Open subtitles corpus, and finally fine-tuned with the in-domain data

\item We fine-tuned all of our models without freezing any layers in the network, since we had sufficient amount of data to train on.

\item We used dropout when fine-tuning with in-domain data, since it is relatively small compared to the UN and Open subtitle data.

\item We trained our final models with an ensemble of the last eight models, where each model was fine-tuned with the in-domain data.

\end{itemize}

Finally we applied system combination over the outputs of best Neural MT and phrase-based systems using MEMT \cite{Heafield-wmt11}. Our efforts were mainly focused towards the AR$\rightarrow$EN TED task. 
In the end we just replicated our best system for the EN$\rightarrow$AR direction and the QED task. For our best Neural MT system, we were unable to use an ensemble in the EN$\rightarrow$AR direction, since we could not train several comparable models to combine.

\vspace{-2mm}
\section{Data Settings and Pre/Post Processing}
\label{sec:datasettings}

We trained our systems using the data made available through IWSLT 2016 campaign. This contained two in-domain data sets TED talks and QED corpus \cite{guzman-sajjad-etal:iwslt13} and two out-domain data sets UN corpus \cite{ZiemskiJP16} and OPUS data \cite{LISON16.947}. The statistics are shown in Table \ref{tab:stats}. For language model we trained using the target side of the parallel corpus and all the available English data from the recent WMT campaign \cite{bojar-EtAl:2016:WMT1}, and GigaWord and OPUS mono corpus for Arabic.


We segmented Arabic data using both MADAMIRA and Farasa. We found MADAMIRA \cite{pasha2014madamira} performed 0.1 BLEU points better than Farasa \cite{abdelali-EtAl:2016:N16-3} (See Table \ref{tab:tok}) and decided to use it for the competition. We tokenized the English side using standard tokenizer of Moses. For English$\rightarrow$Arabic, outputs were detokenized using MADA detokenizer. Before scoring the output, we normalized them and reference translations using the QCRI normalizer \cite{sajjad-etal:iwslt13}. 

\begin{table}
\centering
\small
\begin{tabular}{l| r r r r}
{\bf }   & {\bf TED} & {\bf QED} & {\bf UN} & {\bf OPUS} \\
\hline
Stats    & 240K & 153K & 18.5M & 40M \\
\hline
\end{tabular}
\caption{Number of Sentences in Parallel Data}
\label{tab:stats}
\end{table}

\section{Phrase-based System} 

\subsection{Baseline Settings}

We trained phrase-based Moses system, with the settings described in \cite{birch-etal:2014:IWSLT}: a maximum sentence length of 80, Fast-Aligner for word-alignments \cite{dyer-chahuneau-smith:2013:NAACL-HLT}, an interpolated Kneser-Ney smoothed 5-gram language model \cite{Heafield-kenlm}, 
lexicalized reordering model \cite{galley-manning:2008:EMNLP}, a 5-gram operation sequence model \cite{durrani15:cl}, a 14-gram NNJM model \cite{Devlin_2014_acl}, with the baseline settings described in \cite{joty-etAL:2015:EMNLP}. We used default distortion limit, 100-best translation options, phrase-length of 5, monotone-over-punctuation heuristic, cube-pruning with a limit of 1000 during tuning 5000 during test. We used k-best batch MIRA \cite{cherry-foster:2012:NAACL-HLT} for tuning. 
We used cased BLEU \cite{Papineni:2002:BMA} to measure progress. 



\begin{table}
\centering
\small
\begin{tabular}{l| r r r r r}
\hline
{\bf Segmentation}   & {\bf ted-11} & {\bf ted-12} & {\bf ted-13} & {\bf ted-14} & {\bf Avg} \\
\hline
MADA    & 27.5 & 30.6 & 30.4 & 26.3 & 28.7 \\
Farasa  & 27.4 & 30.3 & 30.2 & 26.4 & 28.6 \\
\hline
\end{tabular}
\caption{MADA versus Farasa Tokenization}
\label{tab:tok}
\end{table}

\vspace{-2mm}
\subsection{Data Selection}
\label{subsec:data-selection}
Due to our experience from previous competitions, we were wary of the fact that simply adding the UN data is harmful for the AR$\rightarrow$ MT system, we therefore selected data through MML filtering \cite{Axelrod_2011_emnlp}. We selected 2.5\%, 3.75\%, 5\%, 10\% and 30\% of the UN data and trained MT pipeline by concatenating the selected data with the in-domain data. We did not include Opus data (40 Million Sentences) and NNJM in these experiments to get the results quickly. Table \ref{tab:ds} shows the results. We found  3.75\% ($\approx$685k sentences) to be the optimal threshold. Alternative to data selection, we tried training in- and out-domain phrase-tables separately and using the out-domain phrase-table only as a back-off. Second last row of Table \ref{tab:ds} shows results. While it gave improvement on top of the baseline system, it was slightly behind MML filtering. 

We then tried to find optimal cut-off on the OPUS data, and selected 20 Million sentences (half of the Opus). Our best systems used 3.75\% of the UN data and half of the Opus data. Adding the selected Opus data gave an average improvement of +1.2 BLEU points. 

\begin{table}
\centering
\small
\begin{tabular}{l| r r r r r}
\hline
{\bf Percentage}   & {\bf ted-11} & {\bf ted-12} & {\bf ted-13} & {\bf ted-14} & {\bf Avg} \\
\hline
ID   & 27.5 & 30.6 & 30.4 & 26.3 & 28.7 \\
2.5\%  & 27.3 & 30.6 & 31.5 & 27.0 & 29.1 \\

3.75\%  & 27.1 & 30.7 & 31.6 & 27.2 & 29.2 \\
5\%  & 27.1 & 30.4 & 31.4 & 27.1 & 29.1 \\
10\%  & 27.0 & 30.2 & 31.5 & 27.2 & 29.0 \\
30\%  & 26.3 & 29.5 & 30.9 & 26.7 & 28.4 \\
Full  & 25.3 & 28.7 & 29.4 & 25.5 & 27.3 \\
\hline
Back-off PT & 27.0 & 30.4 & 31.5 & 27.3 & 29.1 \\
\hline
3.75\%+$\frac{1}{2}$OPUS & 28.2 & 32.4 & 32.3 & 28.6 & 30.4 \\
\hline
\end{tabular}
\caption{Data Selection using MML and Back-off PT}
\label{tab:ds}
\end{table}

\vspace{-2mm}
\subsection{Language Model}

We trained bigger language model by using all the available English data from the recent WMT campaign\footnote{http://www.statmt.org/wmt16/translation-task.html} and target-side of the parallel data. A Kneser-Ney smoothed 5-gram language model was trained on each sub-corpus individually and then interpolated to minimize perplexity on the target part of the monolingual data. We were able to obtain a gain of +0.5 using bigger language model. See Table \ref{tab:bigLM}.

\begin{table}
\centering
\small
\begin{tabular}{l| r r r r r}
\hline
{\bf System}   & {\bf ted-11} & {\bf ted-12} & {\bf ted-13} & {\bf ted-14} & {\bf Avg} \\
\hline
Baseline & 28.2 & 32.4 & 32.3 & 28.6 & 30.4 \\
+bigLM  & 28.3 & 32.8 & 33.2 & 29.2 & 30.9  \\
\hline
\end{tabular}
\caption{Bigger Language Model}
\label{tab:bigLM}
\end{table}

\vspace{-2mm}
\subsection{Interpolation of Operation Sequence Models}

The OSM model has been a regular feature of the phrase-based pipeline \cite{durrani-EtAl:2013:Short} in the competition grade systems. It is a joint sequence translation model which integrates reordering. \cite{durraniEtAl:MT-Summit2015} recently found that an OSM model trained on plain concatenation of data is sub-optimal and can be improved by training OSM models on each domain individually and interpolating them by minimizing perplexity on the in-domain tune-set. Table \ref{tab:IOSM} shows that using interpolated OSM model (OSM$_{i}$) instead of the one trained on plain concatenation (OSM$_{c}$) gives an average improvement of +0.6 BLEU points.

\begin{table}
\centering
\small
\begin{tabular}{l| r r r r r}
\hline
{\bf System}   & {\bf ted-11} & {\bf ted-12} & {\bf ted-13} & {\bf ted-14} & {\bf Avg} \\
\hline
OSM$_{c}$ & 28.3 & 32.8 & 33.2 & 29.2 & 30.9 \\
OSM$_{i}$  & 29.0 & 33.5 & 33.8 & 29.7 & 31.5  \\
\hline
\end{tabular}
\caption{Interpolated Operation Sequence Model}
\label{tab:IOSM}
\end{table}

\vspace{-2mm}
\subsection{NNJM Adaptation}

We also explored the award winning Neural Network Joint Model (NNJM) in our pipeline and tried to adapt it towards the in-domain data. We trained an NNJM models on the UN and Opus data for 25 epochs and then fine-tuned \cite{Luong-Manning:iwslt15} it by running for 25 more epochs on the in-domain data. Because the data is huge, the entire training took 1.5 months of wall-clock time.  Table \ref{tab:nnjm} shows results. The NNJM model gave significant improvement (+0.6) on top of baseline which does not include it. We found fine-tuning method to give slight gains (+0.2) when the baseline model was trained on the Opus data. On the contrary, fine-tuning did not help when the model trained was on UN.


\begin{table}
\centering
\small
\begin{tabular}{l| r r r r r}
\hline
{\bf System}   & {\bf ted-11} & {\bf ted-12} & {\bf ted-13} & {\bf ted-14} & {\bf Avg} \\
\hline
Baseline  & 29.0 & 33.5 & 33.8 & 29.7 & 31.5  \\
+NNJM  & 29.8 & 34.1 & 34.4 & 30.1 & 32.1 \\
+FT(UN)  & 29.7 & 33.8 & 33.9 & 30.2 & 31.9  \\
+FT(OPUS) & {\bf 30.1} & {\bf 34.1} & {\bf 34.6} & {\bf 30.3} & {\bf 32.3}  \\
\hline
\end{tabular}
\caption{Neural Network Joint Model + Different Adaptation Methods}
\label{tab:nnjm}
\end{table}

\vspace{-2mm}
\subsection{Class-based Models}

We explored the use of automatic word clusters in phrase-based models \cite{Durrani-osm-coling14}. We used 50 classes, obtained by running {\tt mkcls}. The clusters ids were included in the phrase-table. We additionally trained in-domain language model using word-classes and interpolated OSM on word-classes. But we only saw very small improvements using word classes. 

\begin{table}
\centering
\small
\begin{tabular}{l| r r r r r}
\hline
{\bf System}   & {\bf ted-11} & {\bf ted-12} & {\bf ted-13} & {\bf ted-14} & {\bf Avg} \\
\hline
Baseline & 30.1 & 34.1 & 34.6 & 30.3 & 32.3  \\
Class-based & 30.3 & 34.2 & 34.7 & 30.4 & 32.4  \\
\hline
\end{tabular}
\caption{Using Word Classes}
\vspace{-5mm}
\label{tab:cb}
\end{table}

\vspace{-2mm}
\subsection{Handling Unknown Words}

We tried to handle OOV words using {\tt drop-oov} and through transliteration \cite{sajjad:acl12,durrani-EtAl:2014:EACL}. The former worked slightly better and was used in the best system. Of course the gains from the two methods are additive because they are addressing different OOVs, but there's no good way to automatically find which word to drop and which one to transliterate.

\begin{table}
\centering
\small
\begin{tabular}{l| r r r r r}
\hline
{\bf System}   & {\bf ted-11} & {\bf ted-12} & {\bf ted-13} & {\bf ted-14} & {\bf Avg} \\
\hline
Baseline & 30.3 & 34.2 & 34.7 & 30.4 & 32.4  \\
{\bf Drop-OOV} & {\bf 30.5} & {\bf 34.2} & {\bf 35.0} & {\bf 30.5} & {\bf 32.6}  \\
Transliteration & 30.4 & 34.2 & 34.7 & 30.6 & 32.5  \\
\hline
\end{tabular}
\caption{Handling OOVs}
\label{tab:oov}
\end{table}

\vspace{-2mm}
\subsection{Final System}

Table \ref{tab:ar-en} shows incremental progress on this Arabic$\rightarrow$English language pair. Our best system included MML selected UN and Opus corpora, big language model, interpolated OSM and fine-tuned NNJM models. We we used {\tt drop-oov} option to handle unknown words.

\begin{table}
\centering
\small
\begin{tabular}{l| r r r r r}
\hline
{\bf System}   & {\bf ted-11} & {\bf ted-12} & {\bf ted-13} & {\bf ted-14} & {\bf Avg} \\
\hline
Baseline  & 27.4 & 30.3 & 30.2 & 26.4 & 28.7 \\
+Selected UN & 27.1 & 30.7 & 31.6 & 27.2 & 29.2 \\
+Selected OPUS & 28.2 & 32.4 & 32.3 & 28.6 & 30.4 \\
+bigLM  & 28.3 & 32.8 & 33.2 & 29.2 & 30.9  \\
+OSM$_{i}$  & 29.0 & 33.5 & 33.8 & 29.7 & 31.5  \\
+NNJM  & 29.8 & 34.1 & 34.4 & 30.1 & 32.1 \\
+FT(OPUS) & 30.1 & 34.1 & 34.6 & 30.3 & 32.3  \\
{\bf Drop-OOV} & {\bf 30.5} & {\bf 34.2} & {\bf 35.0} & {\bf 30.5} & {\bf 32.6}  \\
\hline
\end{tabular}
\caption{Incremental Progress Arabic-to-English System}
\label{tab:ar-en}
\end{table}

\vspace{-2mm}
\subsection{English-to-Arabic Systems}

We did not do detailed experiments for the English$\rightarrow$Arabic direction because of computational limitations, but simply replicated what worked for the Arabic$\rightarrow$English direction. Table \ref{tab:en-ar} shows progress on this language pair. The baseline system (ID) was trained on the the TED data and target side of all the permissible parallel data. In the second row, we added all the parallel data except for the UN. In the third row we additionally added the UN data that we selected in the Arabic$\rightarrow$English direction. Additional parallel data gives an average improvement of +1.4 BLEU point. Then we added an NNJM model trained on in-domain TED data on top of this system to improve it by +0.8. Adding GigaWord and monolingual OPUS data (another 20M Sentences other than the target-side of the parallel data) gave an improvement of +0.3. Finally we replaced the baseline NNJM with the one trained on OPUS data and fine-tuned with the in-domain data to get our best system.

\begin{table}
\centering
\small
\begin{tabular}{l| r r r r r}
\hline
{\bf System}   & {\bf ted-11} & {\bf ted-12} & {\bf ted-13} & {\bf ted-14} & {\bf Avg} \\
\hline
ID & 14.8 & 15.6 & 16.7 & 14.5 & 15.4  \\
+Parallel & 15.5 & 16.4 & 18.2 & 16.3 & 16.6  \\
+MML(UN) & 15.6 & 16.3 & 18.4 & 16.7 & 16.8  \\
+NNJM & 16.5 & 17.4 & 19.2 & 17.4 & 17.6  \\
+bigLM & 16.6 & 17.6 & 20.0 & 17.4 & 17.9  \\
{\bf +NNJM(FT)} & {\bf 16.7} & {\bf 17.9} & {\bf 20.2} & {\bf 17.7} & {\bf 18.1}  \\
\hline
\end{tabular}
\caption{Incremental Progress English-to-Arabic System}
\label{tab:en-ar}
\end{table}

\vspace{-2mm}
\subsection{QED Systems} We simply replicated QED systems by replacing QED corpus to be in-domain data, instead of TED data. We used the same UN data that we selected for our Arabic$\rightarrow$English system, therefore our phrase-tables remain the same. The main changes are caused when training adapted OSM and NNJM models. For NNJM we simply fine tune with QED corpus instead of the TED corpus. For interpolated OSM, we  concatenated TED and QED corpus and build OSM on it, which is then interpolated with the OSM models trained on the selected UN and Opus data. We used IWSLT tuning to get the interpolation weights. This way the OSM sub-model created from TED+QED corpus gets best weights. We also retrained the language model in this similar fashion. We used the tuning weights obtained from our best TED systems and replaced the TED adapted OSM, NNJM and language models with their QED adapted variants.

\vspace{-2mm}
\section{Neural Machine Translation}

\subsection{Pre/Post-processing}

We used a similar pre/post-processing pipeline for Neural MT as
our phrase-based systems (Section \ref{sec:datasettings}), and additionally applied
BPE
\cite{sennrich-haddow-birch:2016:P16-12}
before training them. 
Our BPE models are trained separately for both 
the Arabic and English datasets instead of jointly training them, since the character set differs between the languages. We limited the number of operations to 59,500, as suggested in \cite{sennrich-haddow-birch:2016:P16-12}. We experimented with BPE models trained on the TED data, and on the concatenation of the TED and out-domain data. We did not see any considerable difference in performance between these models. 
Thus we used the BPE model trained on the TED data for the experiments reported in this paper.

\vspace{-2mm}
\subsection{Baseline}

We used default parameters in Nematus to train our systems: a batch size of 80, source and target vocabulary of 50K entries each, 1024 LSTM units, and the embedding layer size of 500. Baseline system were trained using only TED corpus.


\vspace{-2mm}
\subsection{Fine Tuning on Concatenation versus OD}
\label{subsec:model-stacking}

The best phrase based systems are usually trained by concatenating in and out-domain data. On the other hand, deep learning systems are trained on the out-domain data first, and then fine-tuned with in-domain data. We experimented with both strategies. In the interest of time we selected 30\% of the UN data using MML filtering (Table \ref{tab:ds}). We trained two systems, one by concatenating the in-domain data with the selected (30\%) UN data and other just on the selected data. Then we fine-tuned both the models with the in-domain TED data after running them for 3 epochs.
Table \ref{tab:model-stacking} shows that fine-tuning a system trained on out-domain data only, outperforms the system fine-tuned on concatenation.

\begin{table}
  \centering
  \small
  \begin{tabular}{l| r r r r r}
    \hline
    {\bf System}   & {\bf ted-11} & {\bf ted-12} & {\bf ted-13} & {\bf ted-14} & {\bf Avg} \\
    \hline
    30\%+FT(TED) &  27.2 & 31.4 & 30.8 & 27.1 & 29.1 \\
    30\%+TED+FT(TED) &  27.2 & 30.8 & 30.1 & 25.8 & 28.5 \\
    \hline
  \end{tabular}
  \caption{Fine Tuning on Out-domain versus Concatenation -- Models run for 3 epochs}
  \label{tab:model-stacking}
\end{table}

\vspace{-2mm}
\subsection{Fine-tuning Variants and Dropouts}

The default version of Nematus applies fine-tuning by freezing the weights of embedding layer. The intuition behind freezing a layer is to not allow the weights in that layer to change with additional data. This is sometimes useful when we can learn certain layers better from out-domain data. One such layer in our case is the word embedding layer. We tried a variation in which we do not freeze any layer. This latter variant was found to outperform the default setting (See Table \ref{tab:finetune}).


Dropouts are found to be useful in NN training, when the training data is small. We experimented with using dropouts in our experiments, but
did not find any significant difference. Hence we decided to use it only when fine-tuning with the in-domain data (TED/QED), since both of the other datasets (UN and OPUS) were big and did not pose any risk of inducing the problem of overfitting.

\begin{table}
  \centering
  \small
  \begin{tabular}{l| r r r r r}
    \hline
    {\bf System}   & {\bf ted-11} & {\bf ted-12} & {\bf ted-13} & {\bf ted-14} & {\bf Avg} \\
    \hline
    5\% & 25.8 & 29.4 & 29.3 & 25.0 & 27.4 \\
    5\% (Frozen) & 24.7 & 27.7 & 27.4 & 23.9 & 25.9 \\ \hline
    30\% & 27.2 & 30.8 & 30.1 & 25.8 & 28.5 \\
    30\% (Frozen) & 26.5 & 30.4 & 28.9 & 25.0 & 27.7 \\
    \hline
  \end{tabular}
  \caption{Fine-tuning with/without freezing the Embeddings}
  \label{tab:finetune}
\end{table}

\vspace{-2mm}
\subsection{Data Selection}

Since we found data selection useful in the phrase based system, we also trained our neural systems using 5\%, 30\% and 100\% of the UN data.
In these experiments, we concatenated the 5\% and 30\% of the UN data with the in-domain data. To evaluate the most promising models, we trained all of the models until the learning plateaued, and then fine-tuned these models with in-domain data.\footnote{Because we were running experiments in parallel, we were not aware at this point that fine-tuning on out-domain is a better strategy}  The results are shown in in Table \ref{tab:selected-data}. Using only 5\% of the data proved harmful, and the system did not generalize as well as the other models. The model trained on 30\% of the data performed better than the model trained on all the data, by $0.7$ BLEU points. 


In our subsequent experiments we tried to verify if this finding holds when we add the OPUS data. We therefore trained two systems by fine-tuning 30\% selected UN data or full UN data using OPUS. Here the results flipped and the we found that model that used all of the UN data performed better (Compare last two rows in Table \ref{tab:selected-data}). 
Therefore, we decided to focus our efforts on the model trained on the entire UN data for all of the following experiments.

\begin{table}
  \centering
  \small
  \begin{tabular}{l| r r r r r}
    \hline
    {\bf System}   & {\bf ted-11} & {\bf ted-12} & {\bf ted-13} & {\bf ted-14} & {\bf Avg} \\
    \hline
    5\% + FT(TED) & 25.8 & 29.4 & 29.3 & 25.0 & 27.4 \\
    30\% + FT(TED) & 28.4 & 32.7 & 32.9 & 27.8 & 30.4 \\
    Full + FT(TED) & 28.1 & 32.3 & 31.6 & 27.0 & 29.8 \\  \hline
    30\% + FT(OPUS) & 26.1 & 30.6 & 32.5 & 27.1 & 29.1 \\
    Full + FT(OPUS) & \textbf{28.2} & \textbf{31.7} & \textbf{34.3} & \textbf{29.2} & \textbf{30.8} \\
    \hline
  \end{tabular}
  \caption{Data selection}
  \label{tab:selected-data}
\end{table}

\vspace{-2mm}
\subsection{Ensemble}

Ensembling models has shown to give a consistent boost in performance in past best performing systems \cite{sennrich-haddow-birch:2016:WMT}. We therefore experimented with several variations. We found the best performing combination by fine-tuning the last eight models of the UN+OPUS system, and then ensemble these eight fine-tuned models. Performance improvements from the ensemble are shown in Table \ref{tab:ensemble}. The second row shows systems when we fine tune our best system in Table \ref{tab:selected-data} with the in-domain TED data. In the last row we perform ensemble.

\begin{table}
  \centering
  \small
  \begin{tabular}{l| r r r r r}
    \hline
    {\bf System}   & {\bf ted-11} & {\bf ted-12} & {\bf ted-13} & {\bf ted-14} & {\bf Avg} \\
    \hline
    Full + FT(OPUS)     & 28.2 & 31.7 & 34.3 & 29.2 & 30.8 \\
    \hphantom{Full} + FT(TED)     & 31.8 & 36.2 & 36.1 & 30.8 & 33.7 \\
    Ensemble (8) & 32.5 & 37.0 & 37.2 & 31.5 & 34.6 \\
    \hline
  \end{tabular}
  \caption{Ensembling over 8 Fine-tuned Models}
  \label{tab:ensemble}
\end{table}

\vspace{-2mm}
\subsection{Final System}

Our final system was trained by first using all of the UN data. We then continued training on OPUS data. Once learning had plateaued on the OPUS data, we took the last eight models which were very similar in performance, and fine-tuned each of the them using TED data. We then combined these eight fine-tuned models in an ensemble as our final system. The progress is shown in Table \ref{tab:ar-en-progress}. We used the same strategy for the QED systems by fine-tuning the last eight OPUS models with QED data, and combining these in an ensemble. 

\begin{table}
  \centering
  \small
  \begin{tabular}{l| r r r r r}
    \hline
    {\bf System}   & {\bf ted-11} & {\bf ted-12} & {\bf ted-13} & {\bf ted-14} & {\bf Avg} \\
    \hline
    Baseline & 24.0 & 26.4 & 25.2 & 22.4 & 24.5 \\
    UN       & 15.9 & 17.9 & 20.0 & 16.3 & 17.5 \\
	+OPUS    & 28.2 & 31.7 & 34.3 & 29.2 & 30.8 \\
    +TED	 & 31.8 & 36.2 & 36.1 & 30.8 & 33.7 \\
    +Ensemble& \textbf{32.5} & \textbf{37.0} & \textbf{37.2} & \textbf{31.5} & \textbf{34.6} \\
    \hline
  \end{tabular}
  \caption{Arabic-to-English NMT System progress}
  \label{tab:ar-en-progress}
\end{table}

\vspace{-2mm}
\subsection{English-to-Arabic Systems}
We used insights gained from our Arabic-to-English system experiments to train our English$\rightarrow$Arabic systems. 
Our final model for both TED and QED was first trained on all of the UN data, followed by the OPUS data, and finally fine-tuned with the in-domain data. The progress is shown in Table \ref{tab:en-ar-progress}.

\begin{table}
  \centering
  \small
  \begin{tabular}{l| r r r r r}
    \hline
    {\bf System}   & {\bf ted-11} & {\bf ted-12} & {\bf ted-13} & {\bf ted-14} & {\bf Avg} \\
    \hline
    UN       & 9.1 & 9.3 & 11.2 & 9.4 & 9.8 \\
	+OPUS    & 10.8 & 11.2 & 13.4 & 10.9 & 11.6 \\
    +TED	 & \textbf{17.1} & \textbf{18.9} & \textbf{20.1} & \textbf{17.7} & \textbf{18.5} \\
    \hline
  \end{tabular}
  \caption{English-to-Arabic NMT System progress}
  \label{tab:en-ar-progress}
\end{table}

\vspace{-2mm}
\section{System Combination}

We combined hypotheses produced by our best Phrase-based and Neural MT systems. For this purpose we used Multi-Engine MT system, or MEMT \cite{Heafield-wmt11}. The results are shown 
in Table \ref{tab:sysComb}. We did not gain any substantial improvements using system combination. Small improvements were obtained in the Arabic$\rightarrow$English direction baring test-2012. On the contrary significant improvement was obtained only in test-2013 in the English$\rightarrow$Arabic direction. Table \ref{tab:official} shows results on the official test-sets.

\begin{table}
\centering
\small
\begin{tabular}{l| r r r r r}
\hline
{\bf System}   & {\bf ted-11} & {\bf ted-12} & {\bf ted-13} & {\bf ted-14} & {\bf Avg} \\
\hline
\multicolumn{6}{c}{\bf Arabic $\rightarrow$English} \\
\hline
Phrase-based & 30.5 & 34.2 & 35.0 & 30.5 &  32.6  \\
Neural MT & 32.5 &  37.0 & 37.2 & 31.5  & 34.6 \\
\hline
System Comb & 32.8 & 36.5 & 37.4 & 31.7 &  34.6  \\
\hline
\multicolumn{6}{c}{\bf English $\rightarrow$Arabic} \\
\hline
Phrase-based &  16.7 & 17.9 & 20.2 & 17.7 & 18.1  \\

Neural MT & 17.1 & 18.9 & 20.1 & 17.7 & 18.5 \\
\hline
System Comb & 16.8 & 19.1 & 20.7 & 17.6 &  18.6  \\
\hline
\end{tabular}
\caption{Results for System Combination}
\label{tab:sysComb}
\end{table}

\begin{table}
\centering
\small
\begin{tabular}{l| r r r}
\hline
{\bf System}   & {\bf ted-15} & {\bf ted-16} & {\bf qed-15}  \\
\hline
\multicolumn{4}{c}{\bf Arabic $\rightarrow$English} \\
\hline
Primary & 34.1 & 31.8 & 28.1 \\
Contrastive & 33.7 & 31.5 & 28.1 \\
\hline
\multicolumn{4}{c}{\bf English $\rightarrow$Arabic} \\
\hline
Primary & 19.5 & 18.4 & 23.1 \\
Contrastive & 19.5 & 18.1 & 22.9 \\\hline
\end{tabular}
\caption{Results on Official Test Sets}
\label{tab:official}
\end{table}

\vspace{-2mm}
\section{Summary}

We trained a very strong phrase-based system with SOTA features such as OSM, NNJM and  big LM. The system improved greatly by applying domain adaptation. To this end we applied MML-based filtering, interpolated OSM and fine-tuning of NNJM models. Overall, our phrase-based system achieved a gain of 4 BLEU points on top of the baseline system. 
We also applied data selection for training our NMT. However, the NMT systems quickly overfit and did not perform well. Our experiments showed that the NMT system trained on the full UN data performed best, and the final NMT system made use of all the available out-of-domain data. However, the training was performed incrementally, starting with UN data for 50k iterations, fine tuned on OPUS for 25k more iterations and then fine tuned the final model using TED talks for a few iterations. We simply replicated our settings to train QED systems. Finally we applied system combination of the two systems using MEMT.

While it is computationally expensive, we found training a neural MT system much simpler than a competitive phrase-based system, where a lot of sub-components need to be optimized independently to reach the best configuration. On the contrary, an NMT system requires least supervision. Secondly once a neural system is trained, the effort can be easily reused to adapt the system towards another domain, as in this case we simply fine-tuned our UN+OPUS system with the QED corpus. On the contrary, almost all the sub-component of a phrase-based system had to be retrained to adapt the system towards QED corpus.

\bibliographystyle{IEEEtran}
\bibliography{iwslt16}
\end{document}